# Exploiting Structure-from-Motion for Robust Vision-Based Map Matching for Aircraft Surface Movement

Daniel Choate and Jason H. Rife, *Tufts University*

## BIOGRAPHY

**Daniel Choate** is a student in the Mechanical Engineering Ph.D. program at Tufts University in Medford, MA. He works in the Automated Systems and Robotics Laboratory (ASAR) with Dr. Jason Rife. He received his B.S. degree in Mechanical Engineering from Union College in 2023, and his M.S. in Mechanical Engineering and Human-Robot Interaction from Tufts in 2025.

**Jason Rife** is a Professor and Chair of the Department of Mechanical Engineering at Tufts University in Medford, Massachusetts. He directs the Automated Systems and Robotics Laboratory (ASAR), which applies theory and experiment to characterize integrity of autonomous vehicle systems. He received his B.S in Mechanical and Aerospace Engineering from Cornell University and his M.S. and Ph.D. degrees in Mechanical Engineering from Stanford University.

## ABSTRACT

In this paper we introduce a vision-aided navigation (VAN) pipeline designed to support ground navigation of autonomous aircraft. The proposed algorithm combines the computational efficiency of indirect methods with the robustness of direct image-based techniques to enhance solution integrity. The pipeline starts by processing ground images (e.g., acquired by a taxiing aircraft) and relates them via a feature-based structure-from-motion (SfM) solution. A ground plane mosaic is then constructed via homography transforms and matched to satellite imagery using a sum of squares differences (SSD) of intensities. Experimental results reveal that drift within the SfM solution, similar to that observed in dead-reckoning systems, challenges the expected accuracy benefits of map-matching with a wide-baseline ground-plane mosaic. However, the proposed algorithm demonstrates key integrity features, such as the ability to identify registration anomalies and ambiguous matches. These characteristics of the pipeline can mitigate outlier behaviors and contribute toward a robust, certifiable solution for autonomous surface movement of aircraft.

## 1    INTRODUCTION

Present-day commercial aircraft are not designed for automated surface movement. Commercial aircraft rely entirely on human pilots to guide aircraft from the gate to the runway (or vice versa). Automating aircraft movement requires the addition of both high-integrity positioning sensors and electronic actuators for steering. Together these capabilities are needed to measure deviations away from the aircraft's target trajectory and to correct them. In this paper we focus on the position sensor component needed to enable automated surface movement. Specifically, we introduce a novel pipeline for vision-aided navigation (VAN), where the algorithm pipeline is specifically tailored to streamline development of a rigorous integrity case.

Vision is only one of a range of sensor modalities that might be used to enable autonomous surface movement of aircraft. Another key sensor to enable surface movement will be the Global Navigation Satellite System (GNSS) augmented by the Ground-Based Augmentation System (Murphy & Imrich, 2009). This application of GBAS will require new technology development, though, because GBAS provides neither VHF coverage nor meaningful integrity bounds for aircraft on the surface. Surveillance systems might also be adapted for positioning, including secondary radar or multilateration systems (Mario & Rife, 2012). Although these radionavigation options offer benefits, VAN is an attractive option, too, in part because camera-based positioning is strongly analogous to human-pilot vision, on which modern commercial aircraft rely.

The basis of VAN revolves around the 3D reconstruction problem, specifically generating 3D structure from a series of 2D images. Several solutions have been explored to address this problem (Snavely *et al.,* 2010). In the computer vision space, these solutions fall under different umbrellas of structure-from-motion (SfM), simultaneous localization and mapping (SLAM), and visual odometry (VO) (Herrera-Granda *et al.*, 2024). Scene reconstruction can be addressed with a variety of solutions spanning classical geometric methods (Schonberger & Frahm, 2016; Wang *et al.,* 2013), probabilistic approaches (Hornung *et al.* 2013; Pizzoli *et al.,* 2014), and machine learning (ML) approaches. In this paper, we focus on analytical methods rather than ML models like neural networks, as no certification processes yet exist for ML algorithms in the aviation domain.

Recent literature has categorized different reconstruction and SLAM algorithms into two main categories: direct and indirect. Direct approaches utilize full image information without preprocessing steps such as feature extraction or optical flow extraction, while indirect methods utilize such preprocessing steps (Herrera-Granda *et al.*, 2024). While common feature-based indirect methods (Campos, *et al.*, 2021) can be robust and more computationally efficient than direct methods, direct methods offer an important integrity aspect when considering full image and pixel information (Silveria *et al.,* 2008; Engel *et al.,* 2018; Zubizaretta *et al.,* 2020). Direct methods also avoid the issue of incorrect feature correspondence. Our approach utilizes the computational power of an indirect approach (Schonberger & Frahm, 2016) for initial pose generation of input frames, while leveraging a direct method for final map registration. Using a direct method for the final step has the potential to enhance integrity, both by avoiding incorrect feature extraction and correspondence and by providing supplemental information about registration quality.

The airfield environment adds specific challenges, which somewhat differentiate our VAN approach from others tailored for scenarios like agriculture (Bai *et al.,* 2023), roadway navigation (Scaramuzza & Siegwart, 2008; Nourani-Vatani & Borges*,* 2011), and offroad navigation (Gonzalez *et al.,* 2012). For example, in an airport environment the terrain is extremely flat with sparse 3D queues, a problem for terrain-aided approaches (Navon *et al.,* 2025). Also, the environment is highly structured with numerous ground markings (lights, painted lines and numbers); however, those markings are often repeated in multiple locations and therefore not unique. Color queues (Magnusson, 2009; Gupta *et al.,* 2023) are sometimes available; however, the value of color queues is somewhat limited by lighting sensitivity and low visual texture in the airfield environment. The environment is also dynamic, with aircraft and other service vehicles moving on the airfield, so some form of anomaly detection and removal is needed (Li *et al.,* 2019; Choate and Rife, 2024).

In practice, we expect that VAN will be most functional, not as a sole sensor for positioning, but rather in a sensor-fusion suite that also includes an inertial measurement unit (IMU) and, if available, GNSS. The integrity of the fused solution will rely on the individual integrity cases for each positioning sensor, however. As such, the integrity of the VAN solution is paramount, even when integrated into sensor fusion system.

The goal of this paper is to begin developing an algorithm both tuned to the airfield environment and to the integrity demands of a safety-of-life application. With integrity in mind, one key feature is that the algorithm is fully interpretable, with transparency achieved through a reliance on geometric analysis and classical signal processing. Another characteristic is that the algorithm combines different visual-reconstruction modalities (both indirect and direct) for redundancy, where indirect processing creates an SfM mosaic from aircraft video and where the direct processing matches the SfM mosaic to a satellite image serving as a map. (With flat terrain and few trees, the features of the airfield are easily visible when viewed from a high altitude, which makes satellite imagery particularly useful as a reference for airfield navigation.)

In the process of developing a VAN pipeline to support aircraft surface movement our primary contributions involve (i) introducing methods to merge ground and satellite imagery of an airfield and (ii) identifying characteristics of our VAN pipeline that contribute to a strong safety case. To highlight those benefits, we analyze algorithm accuracy (and accuracy limitations). We test the algorithm on a dataset collected in an environment with some characteristics similar to those of an airfield: the football field on the Tufts University campus.

The paper is organized as follows. The next section describes our VAN pipeline and analyzes aspects of algorithm performance. The algorithm is then applied to the representative dataset to characterize accuracy and to identify secondary characteristics of the VAN pipeline that could be exploited to address safety concerns under off-nominal conditions. After a brief overview of key experimental results, the paper concludes with a brief summary.

## 2    METHODOLOGY

The goal of our VAN pipeline is to use a satellite image to geo-locate camera data acquired by a taxiing aircraft. In this paper, we refer to camera data acquired by the taxiing aircraft as *ground images*, to contrast them with the satellite image. Altogether, our VAN pipeline consists of four distinct components that together perform aircraft localization. These four components include (i) SfM that performs scene reconstruction using a set of images acquired at ground level, (ii) ground-plane extraction,

(iii) synthesis of a ground-image mosaic, and (iv) registration of the ground-image mosaic to a satellite image. The intention of the first step is to perform relative localization and scene reconstruction. The second and third step transform the reconstruction for comparison to the satellite image, which occurs in the final step. A visual flow diagram is shown in Figure 1.

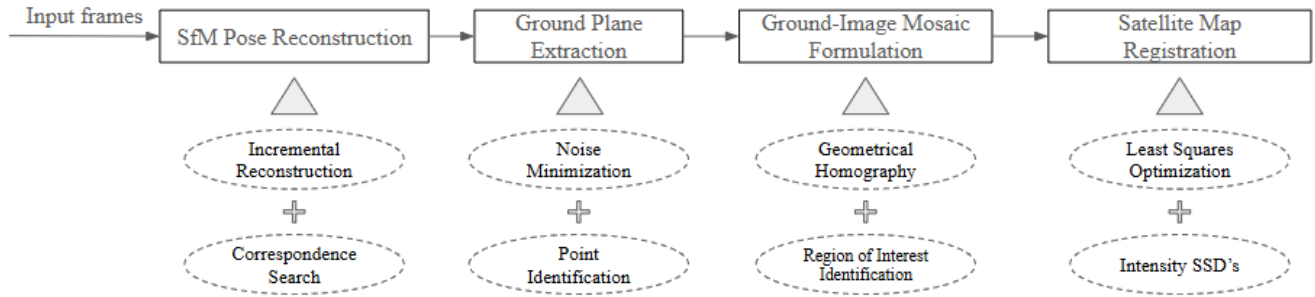

**FIGURE 1**
*Flow diagram representing an overview of our vision-aided navigation (VAN) pipeline.*

For the purposes of algorithm development and initial assessment, we used a surrogate dataset in place of an aviation dataset. Specifically, we acquired imagery at the football stadium on the Tufts University campus. This stadium location features many of the characteristics of an airfield, as the football field is flat, visible from space, and surveyed with distinct and regular markings. A satellite image of the stadium from Google Earth is shown in Figure 2. Our associated ground-image dataset was acquired using an iPhone 12 camera, consisting of a 12MP ultra-wide lens with 120-degree field of view. The ground-image dataset consisted of 142 image frames, snapped individually while walking near the end of the field shown on the right side of Figure 3. The attitude and location of the camera were adjusted between each image frame, to assess the ability of our VAN pipeline to recover the changes in relative camera pose between frames.

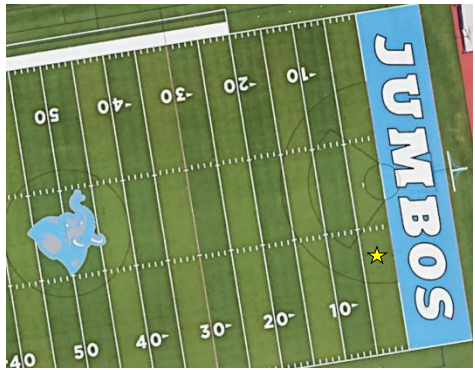

**FIGURE 2**
*Satellite image gathered from Google Earth, with a star indicating the origin used for local image registration*

## 2.1   SfM Scene Reconstruction using Ground Imagery

The first stage of our pipeline performs scene construction using a set of images acquired by the aircraft when on or near the ground, shown by the "SfM Pose Reconstruction" stage, in Figure 1. In this paper, we refer to these images as *ground images*, to contrast them with the satellite image introduced later. The goal of scene reconstruction is twofold: (i) to characterize the planar surface containing the runway or taxiway and (ii) to stitch together multiple images, with the goal of enhancing registration to the satellite-image map.

We implement scene-reconstruction using an open-source SfM algorithm called COLMAP. This approach is well suited to address the research question: what are the limits of vision-only scene reconstruction in our application? SfM runs as a batch process, which searches for the best alignment of all frames in a set. Though visual SLAM methods, which run recursively, are often selected for real-time applications because of their computational efficiency, SLAM methods can struggle to discover relationships between non-sequential images. Thus, SfM is the better choice for an initial development phase focused less on real-time implementation and more on characterizing best-case performance for an analytical vision-only solution.

COLMAP functions by extracting features from each image frame and then matching those features across frames. The algorithm features are extracted using SIFT (scale-invariant feature transform). An iterative bundle-adjustment method is used to transform the estimated relative position and attitude for each frame, to best align the extracted features. The bundle-adjustment process also calibrates the camera intrinsic parameters. The end result is a scene reconstruction that takes the form of a point cloud, where each point represents the triangulated location of a feature. This process is illustrated in Figure 3, which on the left shows 7 of 142 ground images input to COLMAP, and which on the right shows the output point cloud. The point cloud includes rigid frames (visualized as orthogonal red-green-blue vectors) that represent the estimated position and attitude of the camera at the moment each ground image was acquired. Each feature and camera pose are stored using an arbitrary coordinate system and basis, which we identify as the COLMAP frame $C$.

An important aspect of the COLMAP output is that distances can only be determined subject to an unknown scale factor. In other words, the length scale is unknown. This is a standard characteristic of all SfM solutions (Zhang, 2023), because an unaided camera cannot differentiate between a scene and an exact replica of that scene reproduced at a different scale (e.g. in miniature). The unknown scale factor will be resolved later in the VAN pipeline.

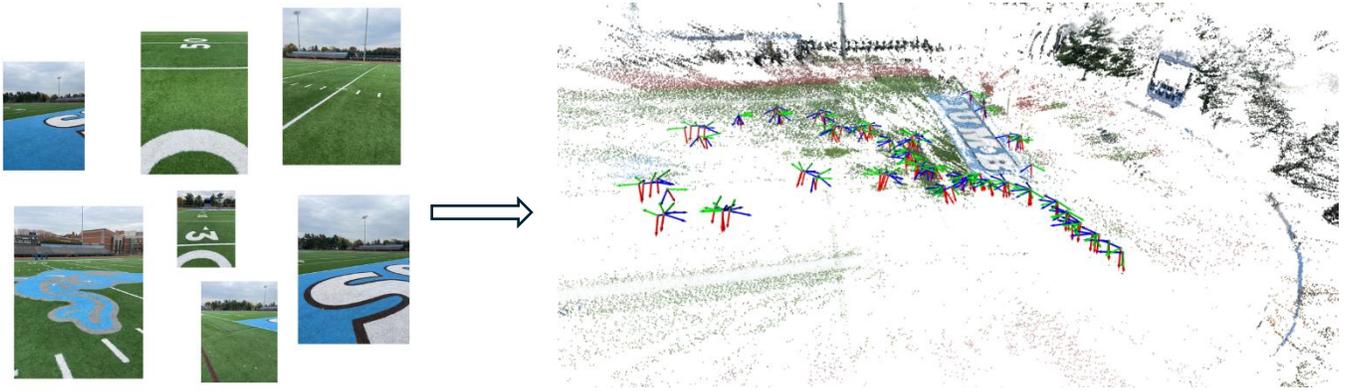

**FIGURE 3**
*Visual representation of SfM solution. Ground images (left) are transformed into a point cloud (right) where the orthogonal red-green-blue vectors indicate the estimated camera pose for each ground image.*

## 2.2   Ground-Plane Extraction

One major benefit of utilizing a SfM solution such as COLMAP, is the ability extract a large planar surface from the point cloud. In our pipeline, we assume this large plane to be the ground. A reference point on the ground plane can then be georeferenced by registering the COLMAP ground plane to a satellite image. In order to characterize a set of points for planarity, we fit an ellipsoid to those points and check that the points are tightly clustered in at least one direction. We use the singular value decomposition (SVD) for this purpose. Let's consider a group of $N$ points from the 3D SfM point cloud, of which at least three are non-coplanar. Each point is described by a vector $\mathbf{a}_i = [x_1 \quad y_1 \quad z_1]$, with coordinates written in terms of frame $C$. The average of these points is then removed to create mean-free vectors $\mathbf{a}'_i$, where

$$\mathbf{a}'_i = \mathbf{a}_i - \left(\frac{1}{N}\sum_{i=1}^{N} \mathbf{a}_i\right). \tag{1}$$

These points locations are concatenated to create the matrix $\mathbf{A} \in \mathbb{R}^{n \times 3}$, where

$$\mathbf{A} = \begin{bmatrix} \mathbf{a}_1 \\ \vdots \\ \mathbf{a}_n \end{bmatrix}. \tag{2}$$

The best fit ellipsoid is computed with the SVD, which finds unitary matrices **U** and **V** and a rectangular diagonal matrix $\Sigma$, where $\Sigma$ describes the semi-axis lengths and **V** their orientations. The SVD computes these matrices such that their product is

$$\mathbf{A} = \mathbf{U}\Sigma\mathbf{V}^T. \qquad (3)$$

The matrix **V** has three unit-length, orthogonal columns, where one of the columns is aligned with the shortest principal axis of the ellipsoid fit to the points. For a planar set of points, this short-axis is perpendicular to the plane, nominally parallel to the gravity direction. (The other two columns lie in the plane.) As such **V** forms a basis for the plane, and it can be viewed as a rotation matrix connecting the arbitrary COLMAP coordinates $C$ to a new coordinate system $G$ that is aligned with the ground plane. In concept, the ellipse centroid and the basis **V** can be used to define an extended plane, to test if other points not in the original set of vectors $\mathbf{a}_i$ are also members of the same plane; in practice, we streamlined this process of automatic plane definition by relying on a human operator to identify regions of interest associated with the ground plane.

## 2.3 Synthesis of Ground-Image Mosaic

Once the ground-plane is extracted from the SfM point cloud, the associated features can, in concept, be registered with a satellite image by rotating those SfM features, translating them laterally, and scaling distances between, so that the SfM ground features align optimally with corresponding satellite-image features. In practice, however, this process is challenging because the features extracted from the ground image do not necessarily work well for matching to the satellite image, a factor which erodes integrity due to high probabilities of both incorrectly extracting and corresponding features. Moreover, the ground features are also relatively low density. To address these issues, we replace the COLMAP-extracted SIFT features with patches drawn from the original ground images. This process avoids defining arbitrary features while also providing high resolution within each patch. Pixels in each patch are all assumed to lie on the ground (a reasonable assumption if the patches are contained within the convex hull of SIFT features on the ground plane). Figure 4 (left side) visualizes patches (red outlines) for each of five ground images. These patches are transformed, via a homography, to produce a composite set of patches viewed from above, as shown in Figure 4 (right side). We refer to this composite image as a mosaic. The mosaic provides a relatively information-rich representation of the COLMAP data, better suited for registration to satellite image data.

In creating the mosaic, we sample the original ground images to match the pixel spacing of the satellite-image map. To do this we must first assume an initial transformation between the ground-referenced coordinates $G$ and the satellite coordinates $S$; this transformation includes four variables: a scale factor $s$, an in-plane (yaw) rotation $\theta$, and two components of in-plane translation described by the measures $t_p$ and $t_q$. In concept, we would then use automation to define the boundaries of a patch in $S$ coordinates for each of a set of ground images, based on the ground features identified for those images, and interpolate the patch intensity values to the new grid. In our streamlined current implementation, we instead leverage human input to define the patches within individual ground images and invert the above process to map interpolated intensity values to the new grid.

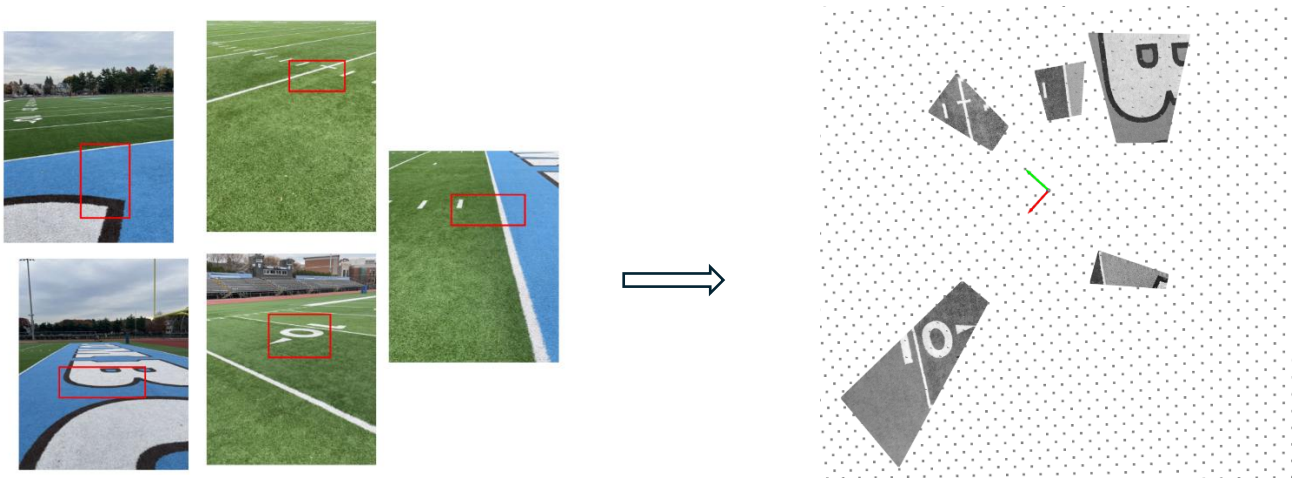

**FIGURE 4**
*Creation of Ground-Image Mosaic. A ground-plane region (red box) is identified in each ground image (left). Using the SfM solution to estimate camera attitude, a homography transform is applied to each patch to build a ground-plane mosaic (right).*

In the process of interpolating patches from individual ground images, we convert color information to grayscale. For this purpose, we use a standard perception-weighted conversion (Zhang, 2023) that computes an intensity scalar $I$ from a color vector **c** with a red component $c_r$, a green component $c_g$, and blue component $c_b$. This conversion is a linear operation with a high weight on the green component and a low weight on the blue component, analogous to the human visual system:

$$I = 0.299R + 0.587G + 0.114B. \qquad (4)$$

The resulting patches are interpolated at the satellite-image scale in the associated satellite-image coordinate system $S$, as shown on the right of Figure 4.

## 2.4 Registration of Ground Mosaic to Satellite Image Map

As shown in Figure 1, the fourth and final block of the pipeline performs a tracking step, iteratively enhancing the initial guess of the transformation parameters $\{s, \theta, t_p, t_q\}$. The initial values of these parameters at a given time step could be identified loosely from other sensors (e.g. from GNSS) or by propagating a former position estimate (e.g. using a first-order hold to propagate velocity from a prior estimated position). The assumption is that the initial values are only a reasonable guess, and not precise. The fourth block of the pipeline performs an estimation process that refines the initial guess by registering the grayscale mosaic and satellite image, such that they align as well as possible. The estimation process is illustrated by Figure 5, which, on the left, shows an initial alignment of patches over the satellite-image map and which, on the right, shows a refinement of the match obtained through our registration process.

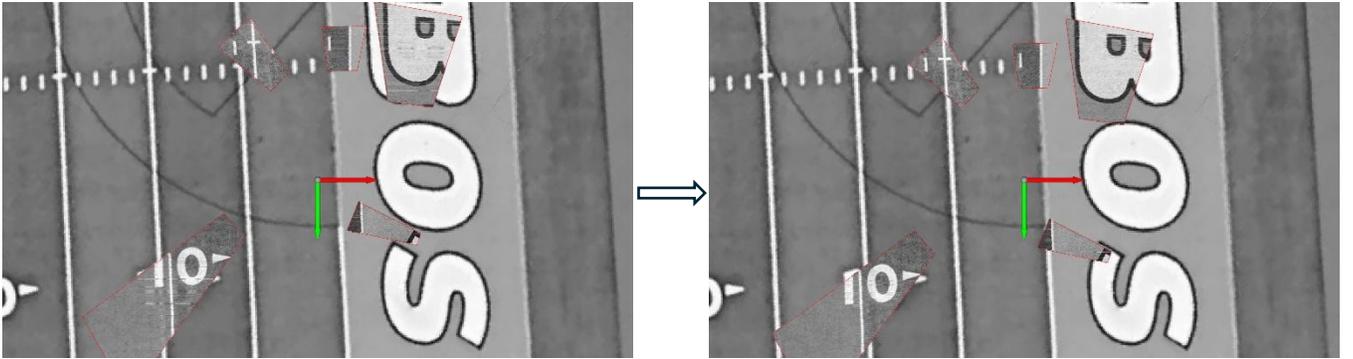

**FIGURE 5**
*Least squares registration for a patch mosaic constructed from five ground images. The image on the left shows the mosaic following an initial 4-DoF guess, while the image on the right shows the patches after convergence.*

Our registration approach can be considered a 'direct' approach, in that it does not require a preprocessing feature-extraction stage (other than for extraction of ground-plane patches, as described above). Importantly, the final registration process (last block of Figure 1) works by directly differencing pixels values between the satellite map and each individual patch. These differences are integrated into a sum of squared differences (SSD) metric, which is minimized when the grayscale values for patches best align with the satellite map. For each ground patch, the alignment that minimizes the SSD function generates a vector $[u_g \ v_g]$ that describes the best alignment of each individual ground patch $g$ to the satellite map.

This SSD function is computed for each ground patch $g$ as a function $SSD_g$ over a space of possible shift vectors defined by $u_g$ and $v_g$:

$$SSD_g(u_g, v_g) = \sum_{(p,q) \in A_g} \sum \left( I_s(p + u_g, q + v_g) - I_g(p, q) \right)^2. \qquad (5)$$

Here $I_s(p,q)$ is the satellite image grayscale value at a pixel location $p$ and $q$, and $I_g(p,q)$ denotes the intensity value for the ground patch $g$ at the same location. The SSD summation combines intensity differences over the set $A_g$ of all pixels in the patch. The process is then repeated for each shift, defined by column displacement $u_g$ and row displacement $v_g$. In our implementation, we considered a minimum of 441 possible integer pixel shifts, covering combinations starting at $u_g \in [-10,10]$ and $v_g \in [-10,10]$.

After finding the best vector shifts for each patch $g$, we use that vector field to refine the parameters $\{s, \theta, t_p, t_q\}$. To this end, consider the coordinate of the patch centroid (the centroid of $A_g$) in two frames. Label the coordinate location of the centroid in the COLMAP ground frame $G$ to be $\bar{\mathbf{x}}_g$, and label the same vector in the satellite frame $S$ to be $\bar{\mathbf{y}}_g$, where the overbar notation is used to indicate the centroid. The centroid vectors are expressed in pixel units, with satellite-image pixels measuring 7.61 cm along each edge. The transformation between the centroid location in $G$ and $S$ is

$$\bar{\mathbf{y}}_g = \mathbf{R}s\bar{\mathbf{x}}_g + \mathbf{t}. \tag{6}$$

Where $\mathbf{R}$ is a rotation matrix between from $G$ to $S$, where $\mathbf{t}$ is the translation between the reference points of each frame, and where s is the scale factor. Expanding $\mathbf{R}$ and $\mathbf{t}$ in terms of the list of states gives:

$$\mathbf{R} = \begin{bmatrix} \cos\theta & -\sin\theta \\ \sin\theta & \cos\theta \end{bmatrix}; \mathbf{t} = \begin{bmatrix} t_p \\ t_q \end{bmatrix} \tag{7}$$

Linearizing (6) about the initial guess, we write the Taylor series expansion as

$$\mathbf{y}_g(\boldsymbol{\alpha}) = \mathbf{h}_g(\boldsymbol{\alpha}) + \mathbf{H}_g \delta\boldsymbol{\alpha} + \cdots \tag{8}$$

where the current guess for the state is the vector $\boldsymbol{\alpha} = [s \quad \theta \quad t_p \quad t_q]^T$. The nonlinear function $\mathbf{h} = [h_p \quad h_q]^T$ can be computed from (6) to have rows

$$h_p = s(\cos\theta\, \bar{x}_{g,p} - \sin\theta\, \bar{x}_{g,q}) \tag{9}$$

$$h_q = s(\sin\theta\, \bar{x}_{g,p} + \cos\theta\, \bar{x}_{g,q}), \tag{10}$$

and the Jacobian $\mathbf{H}_g$ can be computed as

$$\mathbf{H}_g = \begin{bmatrix} \frac{\partial h_p}{\partial s} & \frac{\partial h_p}{\partial \theta} & \frac{\partial h_p}{\partial t_p} & \frac{\partial h_p}{\partial t_q} \\ \frac{\partial h_q}{\partial s} & \frac{\partial h_q}{\partial \theta} & \frac{\partial h_q}{\partial t_p} & \frac{\partial h_q}{\partial t_q} \end{bmatrix}. \tag{11}$$

Assuming higher-order terms in (8) are negligibly small and differencing the equation from the value $\mathbf{y}_g(\boldsymbol{\alpha}_0) = \mathbf{h}_g(\boldsymbol{\alpha}_0)$ at the initial guess $\boldsymbol{\alpha}_0$ gives the linearized equation:

$$\Delta\mathbf{y}_g = \mathbf{H}_g \delta\boldsymbol{\alpha} \tag{12}$$

For any given ground patch $g$, the registration process of (5) creates a vector measurement $\Delta\mathbf{y}_g = [u_g \quad v_g]^T$ that can be substituted into (12) to solve for a state refinement $\delta\boldsymbol{\alpha}$. Given there are four unknowns in $\delta\boldsymbol{\alpha}$ and only two equations in (12), the solution must be obtained simultaneously for at least two patches, to match the number of unknowns and equations, and preferably in a least-squares sense for more than two patches, in order to minimize measurement error. To this end, we define a concatenated measurement vector $\Delta\mathbf{y} \in \mathbb{R}^{2M\times 1}$ and a concatenated Jacobian matrix $\mathbf{H} \in \mathbb{R}^{2M\times 4}$, where $M$ is the number of patches. The concatenated vector $\Delta\mathbf{y}$ stacks each of the individual SSD corrections $\Delta\mathbf{y}_g$ into a tall column; similarly, the concatenated matrix $\mathbf{H}$ stacks each of the Jacobian blocks $\mathbf{H}_g$ into a column. Using these concatenated structures, the state refinement $\delta\boldsymbol{\alpha}$ is computed to be

$$\delta\boldsymbol{\alpha} = (\mathbf{H}^T\mathbf{H})^{-1}\mathbf{H}^T\Delta\mathbf{y}. \tag{13}$$

The process can then be repeated iteratively to convergence, with the initial state estimate updated at each step as

$$\boldsymbol{\alpha}_0 \rightarrow \boldsymbol{\alpha}_0 + \delta\boldsymbol{\alpha} \qquad (14)$$

Curiously, though this least-squares process was developed to register a mosaic to the satellite map, the process can easily be adapted to register an individual ground image to the satellite map if multiple small patches, which we call *micropatches*, are extracted from the same ground image. The micropatches form a mosaic, which allows the SSD vectors to be computed for each micropatch using (5) and a state correction to be computed using (13). We will invoke this single-image alternative to the multi-image implementation for performance comparison in the Results section. An example of registration using micropatches from a single ground image is shown in Figure 6.

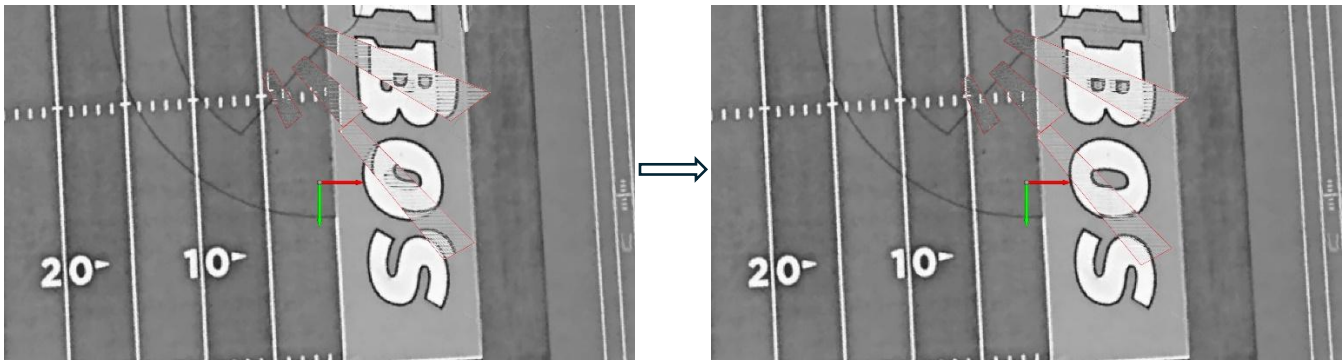

**FIGURE 6**

*Least squares registration for a single ground image using a micropatch mosaic. The image on the left shows the mosaic following an initial 4-DoF guess, while the image on the right shows the patches after convergence.*

## 3      RESULTS AND DISCUSSION

We leveraged the vision-processing pipeline described above to explore two aspects of the problem of registering ground and satellite images in a flat environment. First, we investigated whether constructing a local SfM mosaic can improve ground-satellite registration accuracy. Second, we looked beyond accuracy to identify aspects of the pipeline that contribute to registration integrity, and in particular to cases where intermediate calculations identify defects in the available image data.

### 3.1      Impact of SfM-generated Submap on Registration Accuracy

It seems reasonable to hypothesize that constructing a local SfM mosaic prior to ground-satellite registration might improve accuracy. At least with our implementation, we found no significant improvement in accuracy when registering a multi-image mosaic to the satellite data rather than a single-image mosaic.

As a mechanism of analyzing the potential limits of registration accuracy, we first quantified consistency errors within the SfM mosaic. The SfM mosaic is a batch relative-position estimate and, by extension, is subject to the same drift errors common to other dead-reckoning methods. Translation and rotation errors can stack, leading to growing drift across the SfM point cloud. At the same time, estimation of scale $s$ and yaw rotation $\theta$ are enhanced when patches are separated by a long baseline. Because there is a balance in selecting the length scale of the mosaic, trading off the disadvantages associated with drift and the advantages associated with a larger baseline, we might ask if a "goldilocks" length scale exists that provides the best tradeoff. In practice, we quickly found that drift was too large to register the entire set of ground images to the satellite image. The longest length scale we considered was associated with a set of recent images (as many as five ground images); the smallest length scale was that of a single image (with micropatches extracted from different segments of the ground identified within that image).

To test the quality of the SfM across different mosaic sizes, we defined interpatch error. The interpatch error compares the true distances between patch centroids, as evaluated from the satellite map, to the inferred distances between patch centroids as predicted by the original SfM solution. Due to dead-reckoning drift, the difference between the true and the SfM distances is expected to grow as the interpatch distances increase. To quantify interpatch error, we start by defining the ground-truth vectors $\bar{\mathbf{y}}_i^*$ and $\bar{\mathbf{y}}_j^*$ as the centroids of patches $i$ and $j$ in the coordinates of satellite frame $S$ if they were moved individually to the locations minimizing the SSD function. The asterisk (*) superscript here indicates the optimal SSD solution for the mosaic registration. The ground-truth separation $\boldsymbol{\Delta}_{ij}^*$ is

$$\boldsymbol{\Delta}_{ij}^* = \bar{\mathbf{y}}_i^* - \bar{\mathbf{y}}_j^*. \tag{15}$$

The estimated centroid positions for both patches, after the registration step (13), are $\bar{\mathbf{y}}_i$ and $\bar{\mathbf{y}}_j$. When more than two patches are used in the registration process, the least-squares solution pushes the patches toward the minimum SSD locations, but the mosaic constraints prevent them from matching exactly. As such, the estimated separation

$$\boldsymbol{\Delta}_{ij} = \bar{\mathbf{y}}_i - \bar{\mathbf{y}}_j \tag{16}$$

differs slightly from the ground-truth separation $\boldsymbol{\Delta}_{ij}^*$. We quantify the interpatch error $\boldsymbol{\epsilon}_{ij}$ as the difference between these two:

$$\boldsymbol{\epsilon}_{ij} = \boldsymbol{\Delta}_{ij}^* - \boldsymbol{\Delta}_{ij}. \tag{17}$$

In practice, it is not necessary to compute all of the terms in (15) and (16) individually to obtain $\boldsymbol{\epsilon}_{ij}$. Rather it is sufficient to simply difference the least-squares residuals from the registration step (13), since each residual $\boldsymbol{r}_i = \bar{\mathbf{y}}_i^* - \bar{\mathbf{y}}_i$. By extension, the residuals can be used to compute the interpatch error as $\boldsymbol{\epsilon}_{ij} = \boldsymbol{r}_i - \boldsymbol{r}_j$.

Working with the football-field example, we computed interpatch error for the SfM solution illustrated in Figure 3. First, we worked through the process of building a five-image mosaic using the images and regions-of-interest illustrated in Figure 4. The resulting best-case registration of those five ground images to the satellite image is shown in Figure 5. The corresponding residuals from (13) were used to compute the interpatch errors $\boldsymbol{\epsilon}_{ij}$ for a superset of ten patches (the original five patches plus five more). This process generated a total of 45 interpatch error values. The 45 absolute errors $|\boldsymbol{\epsilon}_{ij}|$ are plotted in Figure 7 as a function of the estimated distance between patches $|\boldsymbol{\Delta}_{ij}^*|$.

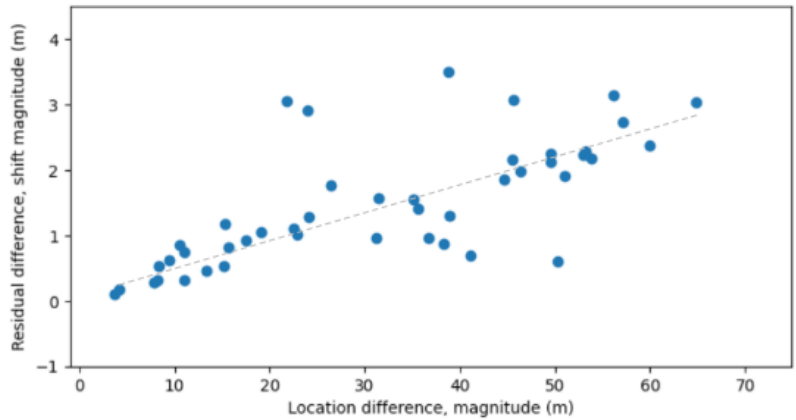

**FIGURE 7**
*Absolute interpatch errors as a function of estimated distance between patches. A trend line indicates the nominal proportionality between errors and patch separation distance.*

The data shown in Figure 7 indicate that, within the larger-scale SfM solution (across the superset of 10 ground images), distortion errors grow proportional to distance. Unfortunately, this linear error-growth trend counteracts the corresponding linear suppression of rotation and scaling errors when the baseline between patches is large. Considering both effects, the key outcome is that there appears to be no theoretical advantage to conducting registration with a larger mosaic baseline. In other words, there does not appear to be a "goldilocks" mosaic size that minimizes error in estimating rotation $\theta$ and scale factor $s$.

To visualize the systematic drift in the SfM solution, we have plotted the patch centroids for the superset of ten patches in Figure 8. The red points represent the transformed SfM and the green points represent ground truth. The five patches used to perform the least-squares registration are located near the lower-right corner of the satellite map, where the red and green points are neatly overlaid. Although the scale, rotation, and translation solution is good for this compact mosaic of five patches, the same scale, rotation, and translation do not describe the other five patches from the larger-scale SfM solution. Internal dead-reckoning drift within the SfM solution affects not only position, but also rotation and scale, so we should not expect these parameters to be entirely consistent throughout the larger-scale SfM. These systematic drifts are certainly visible in Figure 8.

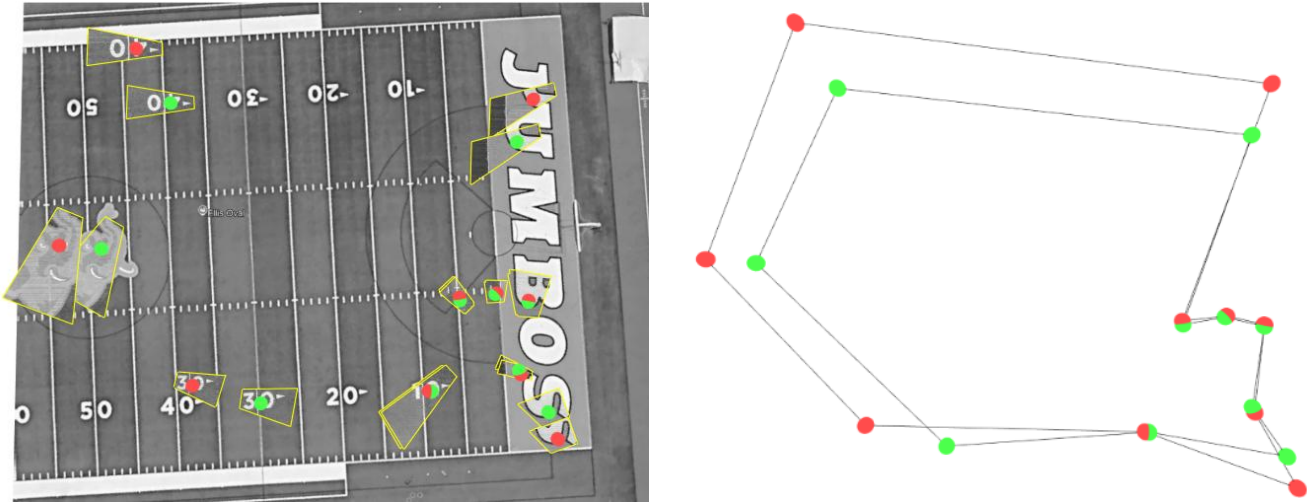

**FIGURE 8**
*A visualization of patches (left) and their centroids (right) after registration of a five-image SfM mosaic to the satellite map. The registration was conducted to align a small cluster of five points in the lower right corner. The red dots represent the estimated patch locations preserving information from the SfM; the green dots indicate the ground truth for individual patches. It is evident that no combination of scaling, rotation, or translation can align all the red and green points.*

Noting that systematic errors grow with distance according to Figure 7, a possible conclusion might be that it is better to perform positioning by registering only the most recent ground image to the satellite map, rather than a multi-image mosaic. However, working with a single image is not a panacea. One practical limitation is that the individual images are aligned to the ground plane using the SfM. Errors in the SfM alignment translate into distortions for any patch, whether for a large patch (for a multi-image mosaic) or a micro-patch (for a single image). Thus, distortions are not resolved simply by switching to single-image registration. A second limitation involves accuracy limitations associated with aligning the single image. A multi-image mosaic contains a larger number of pixel comparisons than a single image and should, by extension, have lower overall uncertainty due to purely random noise.

Perhaps due to alignment issues, systematic errors appear for the single-image mosaic consisting of micropatches, just as they do for a multi-image mosaic. Figure 9 plots inter-micropatch errors as a function of estimated distance between micropatch centroids. The figure was generated for a superset of four micropatches used for registration plus two additional micropatches. Over the superset of 6 micropatches, 15 combinations of two patches were possible. Interpatch errors were computed for each of these pairs. The distances between micropatches are somewhat smaller than the distances seen across multi-image mosaics in Figure 7, because the micropatches are all drawn from a single image. Nonetheless, the trends are very similar between the figures, with the error-versus-distance trend having an approximate slope of 0.035 for the micropatches of the single image (for Figure 9), comparable to the slope of 0.040 for the multi-image mosaic (for Figure 7).

Several mechanisms can contribute to interpatch errors, including pitch/roll misalignment, pixel resolution, and scene changes. Of these mechanisms, the most likely to result in the linearly growing trend in Figure 7 and Figure 9 is the pitch/roll misalignment. We leave further investigation of misalignment effects for future study.

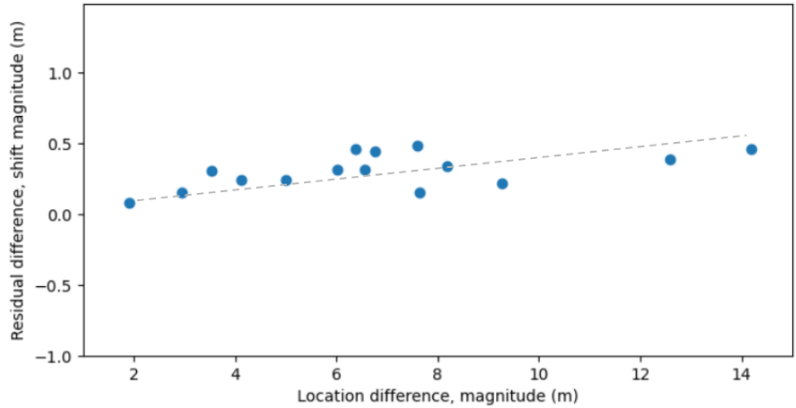

**FIGURE 9**
*Absolute inter-micropatch errors as a function of estimated distance between micropatches. A trend line indicates the nominal proportionality between errors and patch separation distance.*

### 3.2   Towards Navigation Integrity

Several features of the algorithm provided visibility into its integrity. Here, the term *integrity* is used to mean a characterization of off-nominal system performance including rare errors and faults. For an aviation system, hazardously misleading information (HMI) introduced by off-nominal conditions or faults must be guaranteed to occur less often than a specified requirement, in order to ensure system-level safety.

In this section, we describe two elements of our processing pipeline that provide visibility into potential fault conditions. Specifically, we consider (i) the ability of our direct registration method to detect cases of degenerate feature localization and (ii) the ability of a multi-image mosaic to resolve certain types of registration ambiguity.

First, we consider degenerate feature localization. When considering the typical runway and taxiway scenes, environmental markings include extended lines painted along the ground. These markings are useful for lateral localization, but not localization tangent to the line. A high-integrity algorithm should recognize this distinction and treat the line marking different than a more compact marker that supports localization in two axes, such as a painted number. A high-integrity algorithm should also recognize cases when all possible patch-alignments are bad, where the minimum SSD value from (5) can be computed but offers no meaningful information.

As it happens, it is possible to identify degenerate cases like extended markings and poor matches via inspection of the SSD surface (which is a function of the shift parameters $u_g$ and $v_g$). Though we have not developed an automated algorithm to classify the SSD surface for each ground image, we have examined SSD surface properties via human inspection. An example of an extended feature can be seen in Figure 10. The figure shows an ambiguous match for a line feature (yellow outline), where the SSD surface, shown in the middle frame, is trenchlike with low values extending in the same direction as the line. By contrast a more localizable patch (cyan outline) corresponds to the letter "S" in the word "JUMBOS." The SSD surface plot for this cyan patch appears in the rightmost frame of Figure 10. The dark region (low value) appears as a hole in the SSD surface, which is very localizable in two-dimensions. An example of a bad match appears in Figure 11. The figure presents a mis-scaled patch, which is far too magnified to be matched to the satellite image. In this case, the SSD is essentially flat, with low values (blue), all nearly equivalent to the global minimum, spread over an area of roughly 40 square pixels. The flatness of this SSD surface is a clear indicator of a bad match.

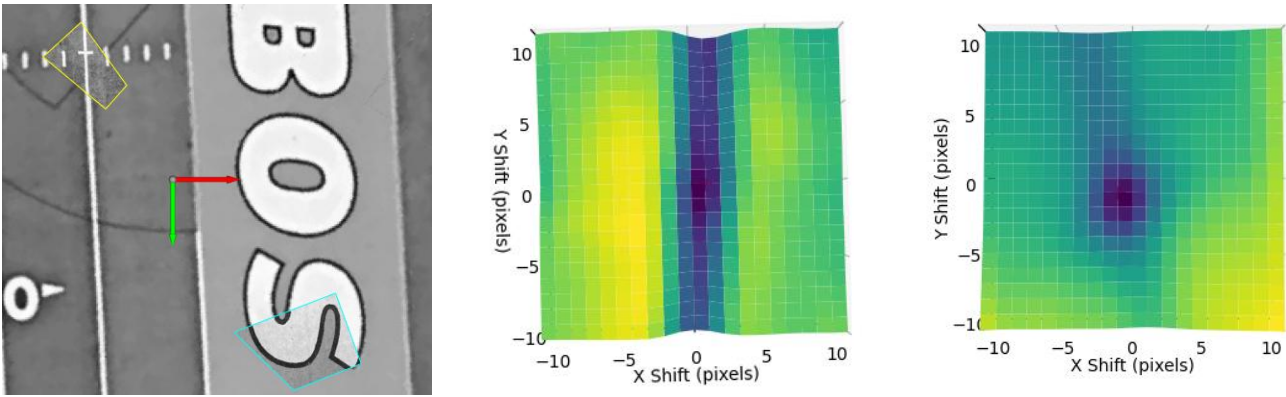

**FIGURE 10**
*Shape of SSD surface. Consider two individually aligned patches (left). The yellow-outlined patch is an extended, largely 1D marking whereas the cyan patch is localizable in 2D. The SSD surface for the 1D case (middle) as viewed from above is very different than the SSD surface for the 2D patch (right). The 1D case appears in the SSD as a long trench with a less-apparent absolute minimum; the 2D case, as a deep hole with a distinct minimum.*

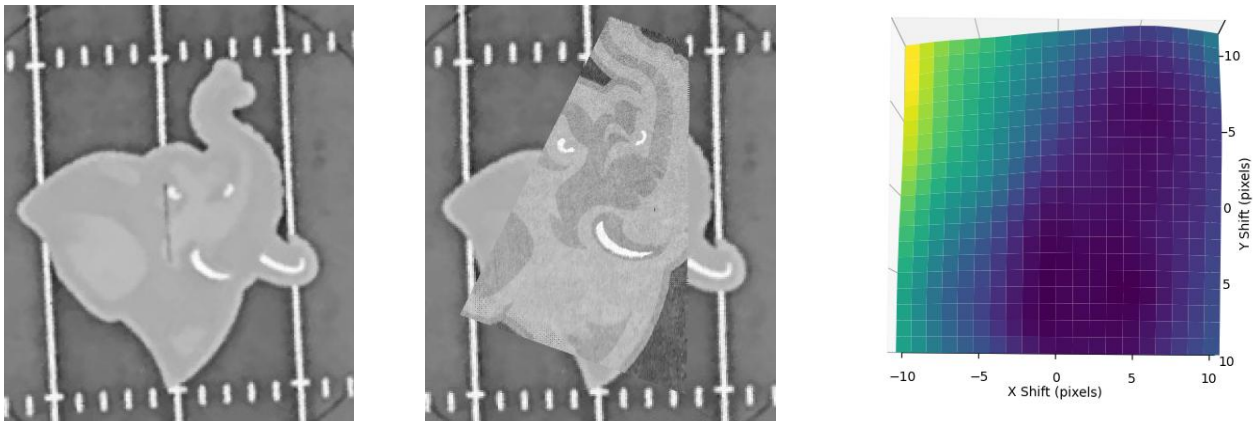

**FIGURE 11**
*Poor quality SSD. The Tufts mascot, Jumbo the elephant, is shown midfield (left). A poorly scaled patch is superposed on top of the satellite image (middle). The resulting SSD surface (right) compares the patch to the satellite map. The SSD surface is very flat, with no crisp minimum, indicating a poor quality registration.*

Second, we consider the case of a registration ambiguity. The ambiguity occurs where a single patch might align with multiple locations in the satellite-image map. For example, consider the repeated pixel pattern seen in Figure 12. Because the SSD surface is only computed over a relatively local area (shifts up to 10 pixels), the SSD surface for an individual patch from Figure 12 would not reveal multiple minima, even though ambiguity is present. As such, the SSD surface alone is insufficient to identify global registration ambiguity. By contrast, the use of a multi-image mosaic can potentially resolve some cases of ambiguity, if the mosaic's length scale is sufficiently large. As an example, consider Figure 13. The figure focuses on a patch taken from a yardage marker. An incorrect, though plausible, registration of the figure is shown on the left side of the figure, surrounded by a red box. The correct global registration is shown in the middle of the figure, surrounded by a green circle. The two cases are nearly indistinguishable, even to a human observer. However, if the registration involves two images linked by SfM, as shown on the right side of the figure, then the ambiguity can be resolved.

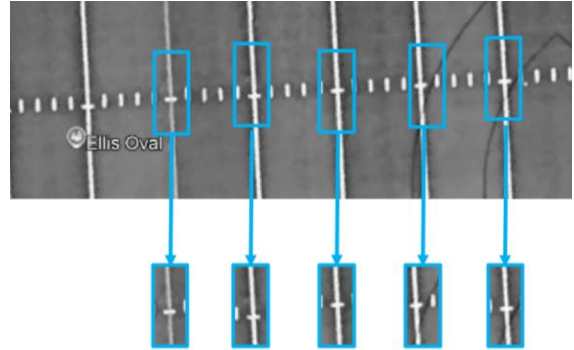

**FIGURE 12**

*Example of an ambiguity in a satellite image of a sports field. A painted cross marker on the field can be registered locally with high accuracy; however, similar cross markers occur at multiple locations along the field.*

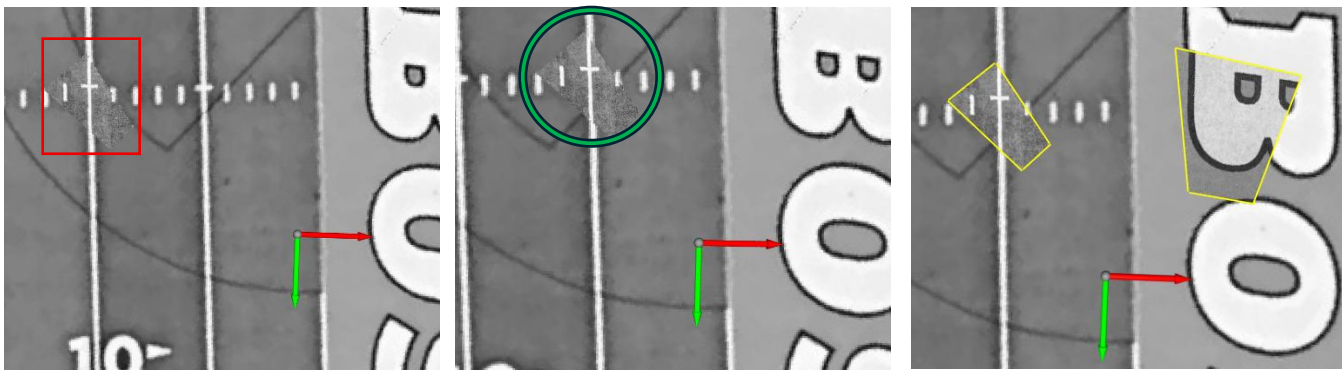

**FIGURE 13**

*An ambiguous registration. The left image shows the patch (boxed in red) registered at the incorrect location. The middle image shows the true location of the patch (circled in green), shifted by one yardage line from the incorrect case. This ambiguity is easily discovered with a multi-image mosaic, as shown on the right.*

### 4    SUMMARY

This paper developed a vision-aided navigation (NAV) pipeline tailored for a relatively flat environment, such as an airfield or sports field. The pipeline was developed from entirely analytical and geometric tools, to ensure its transparency. This transparency is an important factor in meeting the software requirements currently imposed on aviation systems.

We hypothesized that building a local mosaic via SfM would enhance registration accuracy; however, our experimental investigation into registering ground photos to a satellite-image map revealed challenges with SfM-based mosaic generation. Systematic drift within the SfM solution counteracts the potential benefits of a large-baseline mosaic for suppression of random rotation and scale estimation errors. Drift errors grew with distance across the SfM point cloud, at roughly a rate of 4% of interpatch distance, as quantified in Figure 7 and Figure 9. More work is needed to address error growth within the SfM. One possible avenue of improvement may be to use an inertial navigation system (INS) estimates of camera attitude to reduce the attitude-estimation errors inherent to the SfM.

A positive outcome of the study was the identification of integrity benefits inherent to our VAN pipeline. In particular, the pipeline offers tools to characterize degenerate patch registration. A direct comparison of pixel intensity values was used to localize patches from individual ground images to the satellite-image map. The comparison surface, as a function of patch shift over the satellite image, provided visual cues for extended patterns (e.g. long lines cutting across the patch) that provide only one-dimensional localization information. Similarly, the comparison surface appeared visually flat for a case of a badly-scaled patch, where quality registration was not possible. The construction of a multi-image mosaic also offers benefits, in particular for resolving large-scale ambiguities where similar markers are repeated in multiple distinct locations across the satellite image.

ACKNOWLEDGEMENTS

The authors wish to acknowledge and thank the NASA University Leadership Initiative (ULI) Award 80NSSC24M0069 for sponsorship of this work. We also gratefully acknowledge Tufts University, who supported specific aspects of this research. Opinions discussed here are those of the authors and do not necessarily represent those of NASA or other affiliated agencies.